\documentclass{article}
\usepackage{ijcai20}

\usepackage{times}

\usepackage{soul}
\usepackage{url}
\usepackage[hidelinks]{hyperref}
\usepackage[utf8]{inputenc}
\usepackage[small]{caption}
\usepackage{graphicx}
\usepackage{amsmath}
\usepackage{booktabs}
\usepackage{graphicx}
\usepackage{amsfonts}
\usepackage{adjustbox}
\usepackage{enumitem}
\usepackage{romannum}
\usepackage{hyperref}
\usepackage[nomargin,inline,marginclue,draft]{fixme}
\title{Logic Constrained Pointer Networks for Interpretable Textual Similarity}

\author{
Subhadeep Maji$^1$\thanks{Equal contribution.} \footnote{Now at Amazon.}\and
Rohan Kumar$^1$\textsuperscript{*}\and
Manish Bansal$^1$\and
Kalyani Roy$^2$\And
Pawan Goyal$^2$\\
\affiliations
$^1$Flipkart\\
$^2$Indian Institute of Technology, Kharagpur\\
\emails
msubhade@amazon.com,
\{rohankumar, manish.bansal\}@flipkart.com,\\
kroy@iitkgp.ac.in,
pawang@cse.iitkgp.ac.in
}

\begin{document}

\maketitle

\begin{abstract}

Systematically discovering semantic relationships in text is an important and extensively studied area in Natural Language Processing, with various tasks such as entailment, semantic similarity, etc. Decomposability of sentence-level scores via subsequence alignments has been proposed as a way to make models more interpretable. We study the problem of aligning components of sentences leading to an interpretable model for semantic textual similarity. In this paper, we introduce a novel pointer network based model with a sentinel gating function to align constituent chunks, which are represented using BERT. We improve this base model with a loss function to equally penalize misalignments in both sentences, ensuring the alignments are bidirectional. Finally, to guide the network with structured external knowledge, we introduce first-order logic constraints based on ConceptNet and syntactic knowledge. The model achieves an F1 score of 97.73 and 96.32 on the benchmark SemEval datasets for the chunk alignment task, showing large improvements over the existing solutions. Source code is available at \url{https://github.com/manishb89/interpretable_sentence_similarity}
\end{abstract}

\section{Introduction}
Measuring semantic similarity between sentences has been one of the major problems towards text understanding. Many tasks including paraphrase identification~\cite{socher2011dynamic}, text entailment recognition~\cite{heilman2010tree}, etc. also utilize sentence similarity. Clearly, it has attracted a lot of attention in the NLP research community~\cite{shao2017hcti,tai2015improved}. Semantic textual similarity (STS) dataset from SemEval 2012~\cite{agirre2012semeval} has been one of the commonly used benchmark for sentence similarity task, which attempts at measuring the degree of semantic equivalence between two sentences. While recently proposed deep learning methods built on pretrained language models have shown great success for the task \cite{reimers2019sentence}, interpretability and explainability of the final scores remains a concern in general. ~\cite{agirre2015semeval} proposed to formalize interpretable semantic textual similarity (henceforth, iSTS) as an alignment between pairs of segments across the two sentences at SemEval 2016. This is also linguistically well motivated because similarity between sentences has been observed to be decomposable~\cite{sultan2015dls} over segments and the overall similarity is a combined measure of similarity on parts (words, chunks, etc). 
\begin{figure}[t]
    \centering
    \includegraphics[width=\linewidth, keepaspectratio]{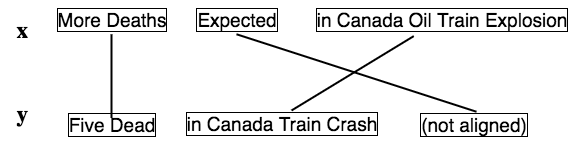}
    \caption{Interpretable sentence similarity defined as alignments between chunks (indicated by connecting lines) of the two sentences. An unaligned chunk is considered aligned to special `not aligned' chunk (denoted by $\phi$ in paper).}
    \label{fig:alignment_example}
\end{figure}
The problem of interpretable semantic textual similarity is to provide an interpretation or explanation of semantic similarity between two texts (usually sentences). We consider chunking to be a preprocessing step and assume that the sentences are already chunked. Not all chunks in the sentences may be aligned and number of aligned chunks gives indication to the overall similarity between the sentences. Figure \ref{fig:alignment_example} provides an illustrative example.

We introduce a novel logic statement constrained gated pointer network model to align constituents of the two sentences, aiding in interpretation of semantic relationships between sentences. Our model uses a pointer network~\cite{vinyals2015pointer} with a sentinel gating function to align the constituent chunks, which are represented using BERT. We improve this base model with a loss function to equally penalize misalignments in both sentences, ensuring bidirectional alignments. Finally, we introduce first-order logic constraints based on ConceptNet as well as syntactic knowledge to aid in training and inference. 

Experiments over two different SemEval datasets indicate that our proposed approach achieves state of the art results on both the datasets. We achieve F1 score of 97.73 on headlines and 96.32 on images, an improvement of 7.8\% and 5.6\% over the previous best results, respectively. Through ablation studies, we also find that the proposed logical constraints help boost the performance on both the datasets. Further, since getting alignments for training the model is costly, we also perform a cross-domain experiment (training on headlines, testing on images and vice-versa) and find that even in this scenario, we achieve F1 scores of 96.16 and 94.80 on these datasets, comprehensively beating state-of-the-art methods. 

\section{Related Work} \label{section:related_work}

Discovering semantic relations in text is a widely studied problem in Natural Language Processing (NLP) with various tasks. One such task is STS which was first introduced as a shared task in \cite{agirre2012semeval}. Several approaches to this task have been developed \cite{bar2012ukp,jimenez2012soft}, with deep learning based methods achieving the most success recently \cite{shao2017hcti,tai2015improved,reimers2019sentence}. 
\cite{shao2017hcti} train a CNN-based sentence encoder on the STS task. \cite{conneau2017supervised} and  \cite{pagliardini-etal-2018-unsupervised} train sentence encoders on alternative similar tasks to aid in learning. Many such approaches make use of an alignment (implicitly or explicitly) between parts of sentences (e.g. \cite{hanig2015exb,sultan2015dls}) with the assumption that composing subsequence alignments can lead to better identification of semantic similarity. Alignments also lend a notion of interpretability of sentence level similarity judgements.

Alignment of parts of sentence pairs is also well studied in domains such as Machine Translation \cite{koehn2003statistical} and Paraphrase Recognition \cite{chang2010discriminative}. The iSTS \cite{agirre2015semeval} shared task focuses on predicting chunk alignments as a means to lend interpretability to STS. \cite{banjade2015nerosim} develop a rule-based alignment system using standard textual features such as parts-of-speech (POS) tags. \cite{kazmi2016inspire} realize a similar rule engine in Answer Set Programming, allowing reordering of rules. \cite{konopik2016uwb} pose chunk alignment as a binary classification task supported with rules. \cite{li2016exploiting} build an alignment model using an integer linear program (ILP), with scoring functions learnt using both alignment and sentence similarity tasks. Chunk alignment is a central piece in this line of work and is also the focus of our work.

Incorporating external information from large labelled or unlabelled corpora into neural models for NLP has been shown to improve task performance. One common way is to use pre-trained word or sentence embeddings. In this work, we experiment with GloVe~\cite{pennington2014glove} and BERT~\cite{devlin-etal-2019-bert} based chunk representations and show ablations in model performance. Another external source of information is in the form of structured knowledge from knowledge bases like ConceptNet~\cite{speer2017conceptnet} and Pharaphrase Database (PPDB)~\cite{pavlick2015ppdb}. Structured knowledge can be incorporated in the form of first-order logic statements without expensive direct supervision. Logic rules provide a declarative language to express such structured knowledge and thus have been used to guide neural networks in several ways recently. 
\cite{towell1990refinement} and \cite{francca2014fast} build networks from a rule set expressed in propositional logic. \cite{hu2016harnessing} make use of teacher-student network formulation to distill logic rules into network parameters. This approach does not restrict models to specialized networks but requires a specialized training procedure. \cite{li2019augmenting} introduce logic statements to constraint labelled neurons (e.g. attention nodes) which can be used in general neural networks. We introduce constraints in the framework which are applicable to the chunk alignment task.
\section{Approach}
\begin{figure}[!htb]
    \centering
    \resizebox{\linewidth}{8cm} {
    \includegraphics[]{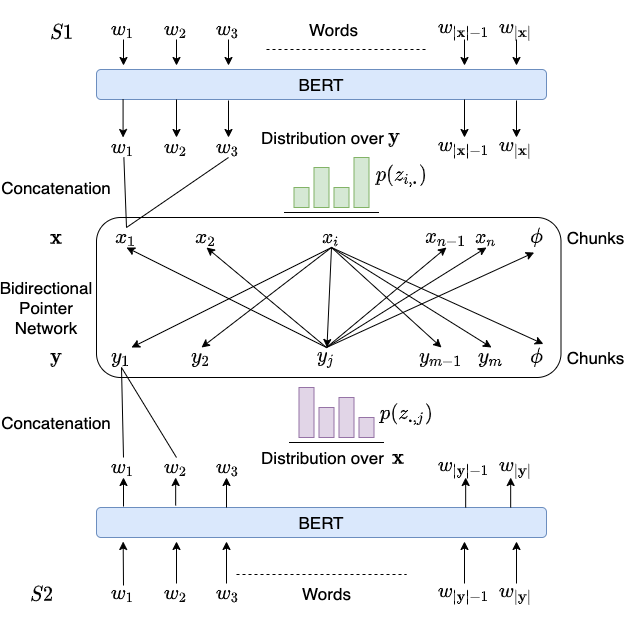}
    }
    \caption{Block level illustration of our Bidirectional Pointer Network with BERT based chunk embeddings. For ease of illustration, we show the non-aligned chunk $\phi$ on both sentences and do not show FOL constraints. }
    \label{fig:model_arch}
\end{figure}

Given two sentences, let us denote these by $\mathbf{x}=(x_1, x_2,\ldots, x_n)$ and $\mathbf{y}=(y_1,y_2,\ldots, y_m)$, where $x_i$ and $y_j$ are chunks (i.e contiguous words) in $\mathbf{x}$ and $\mathbf{y}$, respectively. The problem of iSTS is to predict an alignment between chunks $z_{i,j}\in\{0,1\}$ indicating if $x_i$ and $y_j$ are aligned. We consider a supervised setting in which the ground truth alignment $a_{i,j}\in\{0,1\}$ is specified in the training data. Not all chunks in $\mathbf{x}$ are aligned to a chunk in $\mathbf{y}$ (and vise-versa). We consider these non-aligned chunks to be aligned to a special $\phi$ chunk. Therefore, the problem is to generate an alignment from $\mathbf{x}$ to $\mathbf{y} \cup \{\phi\}$ and from $\mathbf{y}$ to $\mathbf{x} \cup \{\phi\}$. Figure~\ref{fig:model_arch} shows a block level view of our model. Next, we explain each of the individual components in further details.

\subsection{Chunk Representation} \label{subsection:bert_chunk_emb}
Given a chunked sentence, we obtain chunk representation from BERT \cite{devlin-etal-2019-bert} $bert$-$base$-$uncased$ variant by concatenation of contextualized embedding of first and last word of a chunk. We use summation of last 4-layers from BERT to represent a word as higher layers of BERT capture semantic features as suggested in \cite{jawahar2019does}. We use BERT WordPiece model for sub-words representations corresponding to a word which are averaged to realize the word final representation. We also experiment with GloVe where the chunk representation is obtained by averaging the word embeddings, as is common in literature.

\subsection{Alignment with Gated Pointer Networks} \label{subsection:gated_pn}
Extending previous work~\cite{vinyals2015pointer}, we design a pointer network (PN) to model alignment between chunks of sentences $\mathbf{x}$ and $\mathbf{y}$. The alignment between chunks $x_i$ and $y_j$ is modelled as 
\begin{equation} \label{eq:align_pn}
  \theta_{i,j} = v^T f(W_1 x_i + W_2 y_j + W_3 x_i \otimes y_j)  
\end{equation}
where $W_1$, $W_2$, $W_3$ and $v$ are model parameters, $f$ is a non-linearity (we use {\sl tanh}) and $\otimes$ is the Hadamard product. The $W$ matrices are of the same dimension and project chunk embeddings to a lower dimension $d$; $v$ is a $d$-dimension vector. Thus, the PN `points' from chunks in $\mathbf{x}$ to chunks in $\mathbf{y}$. However, we consider the pointers as unidirectional and interpret Equation~\eqref{eq:align_pn} as alignment between $y_j$ and $x_i$ as well. To allow for non-aligned chunks, we consider such chunks to be aligned to a special empty chunk $\phi$ and treat its representation as a parameter of the model. Specifically with a 
little abuse of notation, we model non-aligned chunks $x_i$ and $y_i$ using $b^{\mathbf{x}}_{i,\phi}$ and $b^{\mathbf{y}}_{\phi,i}$, respectively, given by
\begin{align} \label{eq:align_phi}
    b^{\mathbf{x}}_{i,\phi} & = v^T f(W_1 x_i + W_2\phi)  \\ \nonumber
    b^{\mathbf{y}}_{\phi,i} & = v^T f(W_1 \phi + W_2 y_i) 
\end{align}
Using Equations~\eqref{eq:align_pn} and~\eqref{eq:align_phi}, we define a gating function that models $x_i$'s alignment 
to $\mathbf{y}$ and 
$y_i$'s alignment to $\mathbf{x}$, 
\small
\begin{align} \label{eq:overall_align}
    g^{\mathbf{x}}_i & = \sigma (c_1 \max_j (\theta_{i,j} - b^{\mathbf{x}}_{i,\phi}) + c_2) \\ \nonumber
    g^{\mathbf{y}}_i & = \sigma (d_1 \max_j (\theta_{j,i} - b^{\mathbf{y}}_{\phi,i}) + d_2)
\end{align}
\normalsize
where $c_1, c_2$ and $d_1,d_2$ are parameters of the model. Intuitively, $g^{\mathbf{x}}_i$ captures how well $x_i$ is aligned to its best alignment (with high $\theta_{i,j}$) in $\mathbf{y}$ in comparison to $\phi$. Mathematically, $ g^{\mathbf{x}}_i \rightarrow 1$ when $x_i$
is better aligned to some $y_j$ in comparison to $\phi$. Let $z_{i,j} \in \{0,1\}$ indicate if $x_i$ is aligned to $y_j$ ($z_{i,j}=1$). 
Using Equations~\eqref{eq:overall_align} and~\eqref{eq:align_pn}, we model the probability of the event $z_{i,j}$ as, 
\begin{align} \label{eq:gated_pn_align}
    p(z_{i,j} = 1) & \propto g^{\mathbf{x}}_i g^{\mathbf{y}}_j \theta_{i,j} 
\end{align}
Thus, the alignment probability of $x_i$ and $y_j$ is proportional to product of $x_i$ being aligned to some chunk in $\mathbf{y}$, $y_j$ being aligned 
to some chunk in $\mathbf{x}$ and $x_i$ being aligned to $y_j$.

The probability of non-alignment of $x_i$ (aligned to $\phi$) 
is proportional to $1-g^{\mathbf{x}}_i$ and probability of non-alignment of $y_j$ is proportional to $1-g^{\mathbf{y}}_j$. Thus
\begin{align} \label{eq:gate_pn_unalign}
    p(z_{i,\phi} = 1 ) & \propto (1 - g^{\mathbf{x}}_i) \\ \nonumber
    p(z_{\phi,j} = 1 ) & \propto (1 - g^{\mathbf{y}}_j)
\end{align}

Equations~\eqref{eq:gated_pn_align} and ~\eqref{eq:gate_pn_unalign} are appropriately normalised by passing them through a softmax layer. Mathematically, 
\small
\begin{align} \label{eq:nomalization_pn}
    & \sum_{j=1}^{m} p(z_{i,j} =1) + p(z_{i,\phi} = 1)  = 1 \\ \nonumber
    & p(z_{i, j} = 1) = \text{softmax}([ g^{\mathbf{x}}_i g^{\mathbf{y}}_j \theta_{i,j} ; (1 - g^{\mathbf{x}}_i)])
\end{align}
\normalsize
where $[\boldsymbol{\cdot};\boldsymbol{\cdot}]$ indicates concatenation. 

Therefore, the alignment is modeled for every index $i$ in $\mathbf{x}$ because $p(z_{i,\boldsymbol{\cdot}})$ is a distribution for every $i$. The model is trained by minimizing the categorical cross-entropy loss between correct ($a_{i,j}$) and predicted alignment ($z_{i,j}$). However, this loss alone fails to capture the non-alignment for chunks in $\mathbf{y}$ because the PN `points' from $\mathbf{x}$ to $\mathbf{y}$ and therefore cannot model $\phi$ aligning to multiple $y_j$'s. We address this separately by binary cross-entropy loss between $p_{\phi, j}$ and non-aligned chunks in $\mathbf{y}$ indicated by $a_{\phi, j}$. The combined loss is addition of these two losses, 
\small
\begin{align} \label{eq:overall_loss_1}
    & - \frac{C_1}{n} \sum_{i=1}^{n} \left(\sum_{j=1}^{m} a_{i,j} \log(p_{i,j}) + a_{i, \phi} \log(p_{i,\phi})\right) \\ \nonumber
    & + C_2 \left( - \sum_{i=1}^{n} 
    a_{\phi, i} \log(p_{\phi, i}) + (1-a_{\phi, i}) \log(1-p_{\phi, i}) \right )
\end{align}
\normalsize
where $C_1,C_2$ are positive hyperparameters capturing the relative cost sensitivity between the two loss functions. 

\subsection{Improving Alignments with Bidirectionality} \label{subsection:double_stochastic}
\begin{figure}[!htb]
    \centering
    \includegraphics[width=\linewidth, keepaspectratio]{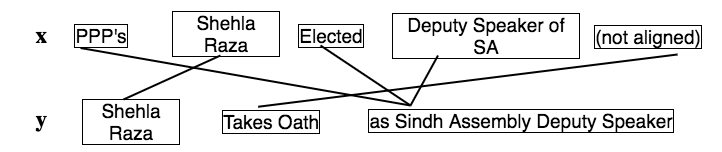}
    \caption{Alignment as obtained by gated PN in Section~\ref{subsection:gated_pn}. Many-to-one alignment obtained by the model for a chunk in $\mathbf{y}$ is incorrect. This is a because it only models $\mathbf{x}$ to $\mathbf{y}$ alignment as a distribution.}
    \label{fig:double_stochastic_single}
\end{figure}

For a pair of similar sentences, usually a certain part of a sentence is semantically similar to only a particular part of the other sentence. Thus, the alignment is usually one-to-one between chunks in the two sentences. 
We observed this phenomenon in SemEval 2016 dataset on iSTS, where 
approximately $85\%$ of alignments between the chunks in the sentence pair are one-to-one. 
However, the gated PN formulation we developed in Section~\ref{subsection:gated_pn} falls short in modelling one-to-one alignments. This is because while alignment for every chunk $x_i$ in $\mathbf{x}$ is modelled explicitly as a distribution $z_{i,\boldsymbol{\cdot}}$, thus encouraging a hard alignment to a chunk $y_j$ (because of the cross-entropy loss), the alignments for $y_j$ are unconstrained (e.g the model might easily choose to align two chunks $x_i$ and $x_k$ to $y_j$). Therefore, the one-to-one alignment for $y_j$'s is violated because it is not being modelled, and is implicitly derived from  $z_{\boldsymbol{\cdot},j}$'s. We illustrate this in Figure~\ref{fig:double_stochastic_single} with an example from the SemEval dataset.
\begin{figure}[!htb]
    \centering
    \includegraphics[width=\linewidth, keepaspectratio]{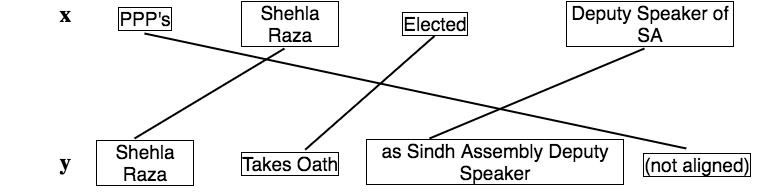}
    \caption{Enforcing bidirectionality on gated PN improves the alignment. The alignment is one-to-one and corresponds 
    to the ground truth alignment.}
    \label{fig:double_stochastic_golden}
\end{figure}

We address this shortcoming of our gated PN by ensuring bidirectionality, i.e.,both $z_{i,\boldsymbol{\cdot}}$ and $z_{\boldsymbol{\cdot}, j}$ are distributions for all $i,j$. Moreover, 
$x_i$'s alignment to $y_j$ and $y_j$'s alignment to $x_i$ is modeled by the same event $z_{i,j}$ with probability $p_{i,j}$. Alignment between two sentences can be viewed as solution to an optimal transport problem where the transportation cost is defined by the activations in Equations~\eqref{eq:gated_pn_align} and~\eqref{eq:gate_pn_unalign}. This results in a combinatorial optimization and is difficult to solve in an end-to-end manner with a neural network. 

We propose an approximate solution to this problem using Sinkhorn distance~\cite{cuturi2013sinkhorn} which is an entropy regularized approximation of optimal transportation problem. Mathematically, Sinkhorn distance is a solution $\mathbf{p}$ to,
\small
\begin{align} \label{eq:sinkhorn_opt}
& \min_{\mathbf{p}} \sum_{i,j \cup \{\phi\}} p_{i,j} C_{i,j} - \lambda H(\mathbf{p}) \\ \nonumber
\text{s.t} & \quad \forall i  \sum_{j \cup \{\phi\}} p_{i,j} = 1 \quad  
 \forall j \sum_{i \cup \{\phi\}} p_{i,j} = 1  
\end{align}
\normalsize
where $\mathbf{p}$ is a distribution, $H(\mathbf{p})$ is the entropy of $\mathbf{p}$, $\mathbf{C}$ is a transportation cost matrix and 
$\lambda$ is the strength of the entropy regularization. 
The constraints in Equation~\eqref{eq:sinkhorn_opt} ensure that $\mathbf{p}$ is a distribution over alignments from $\mathbf{x}$ to $\mathbf{y}$ and from $\mathbf{y}$ to $\mathbf{x}$. We define the cost matrix $\mathbf{C}$ as,
\small
\begin{align*}
    C_{i,j} = - g^{\mathbf{x}}_i g^{\mathbf{y}}_j \theta_{i,j} \\ \nonumber
    C_{i,\phi} = - (1 - g^{\mathbf{x}}_i) \\ \nonumber
    C_{\phi,j} = - (1 - g^{\mathbf{y}}_j)
\end{align*}
\normalsize
In transportation problems, the cost matrix is defined as non-negative. To ensure non-negativity of the cost matrix we subtract the minimum of each row from the entries of the matrix to define, $C_{i,j} = C_{i,j} - \min_{j} C_{i,j} \quad \forall i$. We use Cuturi's iterative row and column normalization algorithm~\cite{cuturi2013sinkhorn} to solve Equation~\eqref{eq:sinkhorn_opt}.
To ensure convergence of Cuturi's algorithm~\cite{cuturi2013sinkhorn}, a small constant $\epsilon$ is added to every entry of the cost matrix. 
The parameters of the model are learnt in an end-to-end manner. The backward pass to compute gradients differentiates through Cuturi's algorithm. This is achievable with automatic differentiation because the algorithm consists of only multiplication and division steps over 
differentiable terms. 

Now both $p_{i,\boldsymbol{\cdot}}$ and  $p_{\boldsymbol{\cdot}, j}$ form distributions and the modified formulation can be viewed as a bidirectional PN which points from both $\mathbf{x}$ to $\mathbf{y}
$ and from $\mathbf{y}$ to $\mathbf{x}$. The loss function is categorical cross-entropy on $\mathbf{p}$. The two-way alignment distribution encourages one-to-one alignments for chunks in both $\mathbf{x}$ and $\mathbf{y}$,
\small 
\begin{align} \label{eq:overall_loss_2}
    &  2 \left(- \sum_{i=1}^{n} \sum_{j=1}^{m} a_{i,j} \log(p_{i,j})\right) \\ \nonumber
    & + \left(- \sum_{i=1}^{n} a_{i, \phi} \log(p_{i,\phi})\right) + \left(-\sum_{j=1}^{m} a_{\phi, j} \log(p_{\phi, j})\right)
\end{align}
\normalsize
Equation~\eqref{eq:overall_loss_2} can be viewed as summation of categorical cross-entropy of distributions $p_{i, \boldsymbol{\cdot}}$ and $p_{\boldsymbol{\cdot}, j}$. Figure~\ref{fig:double_stochastic_golden} illustrates the improvement in alignment on the example 
given in Figure~\ref{fig:double_stochastic_single}.

\subsection{Side-supervision with FOL} \label{subsection:fol_constraints}
Guiding neural networks with structured external knowledge has been shown to improve predictive performance by complementing powerful data-driven learning. This knowledge provides a mechanism to regulate the learning process by encoding human intention without expensive supervision. A number of techniques have been proposed to incorporate declarative logic statements into networks, such as data augmentation~\cite{collobert2011natural}, knowledge distillation~\cite{hu2016harnessing}, etc. \cite{li2019augmenting} propose a promising technique to augment existing networks by constraining activations of named neurons (e.g. attention layer) using declarative rules. Motivated by their approach, in this work, we propose two intuitive rules for the chunk alignment task,
\begin{itemize}[leftmargin = *]
    \item \textbf{R1}: Two chunks should be aligned if they are related by any of the relations: Synonym, Antonym, IsA, SimilarTo, RelatedTo, DistinctFrom or FormOf. The relational information to realize this rule is obtained from ConceptNet.
    \item \textbf{R2}: Two chunks should be aligned if they are syntactically similar. This similarity is defined based on overlap of parts-of-speech tags of their ancestor or children nodes in dependency parse trees of the two sentences.
\end{itemize}
These rules are based on structured knowledge and are expressed as declarative statements. We define predicates $Rel_{i,j}$ and $SynSim_{i,j}$ to indicate whether chunk $x_i$ is related to chunk $y_j$ according to \textbf{R1} or \textbf{R2}, respectively. Let
$A_{i,j}$ denote the model decision that chunk $x_i$ is related to chunk $y_j$. We use the following constraints:
\begin{align} \label{eq:fol_constraint}
    \small
    \forall i,j \in W, Rel_{i,j} \rightarrow A_{i,j}  \\ \nonumber
    \forall i,j \in W, SynSim_{i,j} \rightarrow A_{i,j}
    \normalsize
\end{align}

For rule \textbf{R1}, we experiment with two resources, ConceptNet and the paraphrase database (PPDB). In the ConceptNet knowledge base permitted relation types are [Synonym, Antonym, IsA, SimilarTo, RelatedTo, DistinctFrom, FormOf] which closely correspond to the alignment categories specified in the task \cite{agirre2015semeval}. We also experiment with PPDB since a large number of alignments are paraphrases. As complete chunks may not always be available, we evaluate using bigram and unigram alignments from these sources as indicators of chunk alignments. We ignore non-content words by restricting the alignment to content words $W$, identified as having POS tags [ADJ, ADV, INTJ, NOUN, PROPN, VERB, NUM].

We next discuss the details of syntactic similarity measure which constitutes \textbf{R2}. Using the dependency tree of $\mathbf{x}$ and $\mathbf{y}$ obtained from Spacy \cite{spacy2}, we define syntactic similarity measure between two words $w_1 \in x_i$ and $w_2 \in y_j$ as an average of (\romannumeral 1). Jaccard similarity between parts-of-speech (POS) tags of ancestor nodes of $w_1$ and $w_2$, 
(\romannumeral 2). Jaccard similarity between POS tags of children nodes of $w_1$ and $w_2$ and 
(\romannumeral 3). Boolean to indicate if both $w_1$ and $w_2$ are roots of the corresponding dependency trees. The syntactic similarity between $x_i$ and $y_j$ is defined as the average syntactic similarity between 
words in $x_i$ to its best aligned word (maximum similarity) in $y_j$. A high value of syntactic similarity between $x_i,y_j$
indicates a possible alignment. 

Using the logic statements from Equation~\eqref{eq:fol_constraint} we constraint the pointer network decisions by adding a positive constant $m_{i,j}$ to the activations in Equation~\eqref{eq:align_pn} if either $\textbf{R1}$ or $\textbf{R2}$ is true on $(i,j)$ (i.e. a rule aligns $x_i$ and $y_j$), 
\begin{align} \label{eq:theta_prime}
\small
\theta'_{i,j} = \theta_{i,j} + \rho m_{i,j}
\end{align}
where $m_{i,j}$ captures whether the antecedent $Rel_{i,j}$ or $SynSim_{i,j}$ or both are true.
From an implementation perspective, we unroll the FOL statement to propositional statements for all examples.
The modified alignment strength $\theta'$  in Equation~\eqref{eq:theta_prime} replaces $\theta$
and affects both the gating functions in Equation~\eqref{eq:overall_align} and combined alignment probability in Equation~\eqref{eq:gated_pn_align}. This saturates the probability of the event $z_{i,j}$ to $1$ constraining the network output towards rule predictions. The importance of FOL rules is controlled using the $\rho$ hyperparameter.


\begin{table}[!htb]
    \centering
  \begin{tabular}{ ll }
  \hline
  Hyperparameter & Range\\
  \midrule
  $\rho$ & $[0, 1,2,4]$ \\
  PN dimension($d$) & $[100, 150, 200, 768]$ \\
  \hline
\end{tabular}
\caption{Hyperparameter configurations. The $\rho$ value of $0$ indicates a model without constraints. The PN dimension of $768$ is valid only for BERT chunk based representation models (M3 \& M4).} 
\label{table:hyperparameter_configurations}
\end{table}
\section{Experiments} \label{section:experiments}
We compare the proposed model against existing work on iSTS. In line with much of previous research on this task, we report experimental results on the SemEval 2016 interpretable textual similarity dataset.
\subsection{Dataset Description} \label{subsection:dataset}
We use SemEval 2016 Task 2 dataset for interpretable semantic textual similarity~\cite{agirre2016semeval}. It consists of examples from two domains; News Headlines and Flickr Images. In both domains, there are $1,125$ sentence pairs, with a 2:1 split between train and test sets. Each sentence pair is annotated with the alignments between the chunks, similar to example in Figure~\ref{fig:alignment_example}. In this work, we focus on predicting the alignment between the chunks. Additionally, the chunking of the sentences is pre-specified in both train and test examples, therefore we directly use the chunking in this work. For more detailed specifications, please refer to ~\cite{agirre2016semeval}.         
\subsection{Baselines \& Models} \label{subsection:baselines}

We compare against the best task submissions on the two datasets as well as a followup work achieving SOTA results. Inspire~\cite{kazmi2016inspire} is a rule-based alignment system extending the earlier NeRoSim~\cite{banjade2015nerosim} rule engine. It introduces Answer Set Programming to build an extended rule set. This model was the winning entry of the SemEval task on the News Headlines dataset. UWB~\cite{konopik2016uwb} pose alignment as binary classification between all possible chunk pairs using lexical, syntactic, semantic and WordNet-based features, with impossible alignments handled via rules. This model was the winning entry of the SemEval task on the Flickr Images dataset. Lastly, we compare against \cite{li2016exploiting} which models chunk alignment using an ILP. Scoring functions for the ILP are calculated using a structured loss to penalize alignments far-off from ground truth along with two additional terms for chunk and sentence similarities. To the best of our knowledge, currently this method achieves SOTA results for the iSTS task on the SemEval dataset.

To thoroughly investigate the proposed approach, we report the experimental results across four different configurations of the proposed model, as described in Table \ref{table:alignment_results}. These settings capture the relative merit of each of the model components. The evaluation metric is $F1$ measure as per the SemEval task description.

\begin{table}[t!]
\begin{adjustbox}{totalheight=1.4\height, width=\linewidth}
  \begin{tabular}{ lllllll }
  \hline
  \multicolumn{1}{c}{Model}&
  \multicolumn{4}{c}{Configurations}& 
  \multicolumn{2}{c}{Dataset} \\
  \hline
  & BERT & Glove & FOL & Bidirectional & Headlines & Images\\
  & & & Constraints & PN  \\ 
  \midrule
  Inspire & - & - & - & - & 81.94 & 86.7 \\
  UWB & - & - & - & - & 89.87 & 89.37  \\
  ILP & - & - & - & - & 92.57 & 87.38 \\
  \hline
  M1 & & $\checkmark$ & $\times$ & $\times$ & 89.7 & 88.34 \\
  M2 & & $\checkmark$ & $\times$ & $\checkmark$ & 91.48 & 90.88 \\
  M3 & $\checkmark$ & & $\times$ & $\checkmark$ & 96.63 & 93.81 \\
  M4 & $\checkmark$ & & $\checkmark$ & $\checkmark$ & \textbf{97.73} & \textbf{96.32} \\
  \hline
\end{tabular}
\end{adjustbox}
\caption{Average F1 score of alignments on the test set. The results for the models M1 to M4 are on the best hyperparameter configuration for each. Our best model M4 is better than the existing SOTA method by $6.7\%$ on average across the datasets.}
\label{table:alignment_results}
\end{table}

\subsection{Evaluation} \label{subsection:results}
We report F1 measure across the model ablations and baselines discussed in Section~\ref{subsection:baselines}. The results for the baselines are reported from the corresponding papers owing to unavailability of the their code bases. We trained each of the model configurations on train part of the SemEval dataset and did hyperparamter tuning on training set F1. We fixed the entropy regularization strength $\lambda$ to $0.6$ across all experiments and changing it had little effect on results. The embedding dimension for Glove 
based representations was $300$ and for BERT was $768$. To avoid over-fitting, we employed early stopping using training set F1 as a metric and stop the training if training set F1 does not improve over $5$ successive epochs. In Table~\ref{table:alignment_results}, we report the average test set F1 over $3$ runs for each model configurations (corresponding to the best hyperparameter setting) on news headlines and images dataset. For example, the best identified hyperparameter configuration corresponding 
to M4 are; $\rho=2$ and PN dimension of $100$ for experiments on headlines dataset and $\rho=2$ and PN dimension of $150$ 
for experiments on images dataset. The results for M4 and M3 in comparison to M2 and M1 indicate that BERT based chunk embeddings are superior to Glove based representation. The constraints improve the performance of the BERT based model as highlighted by superior performance of M4 over M3. The improvements using constraints are much more visible on the images dataset. Both BERT chunk representation based models are significantly better than SOTA results~\cite{li2016exploiting,konopik2016uwb}. We see an improvement of 5.6\% on News headlines dataset and 7.8\% on Flickr images dataset for M4 over SOTA results. The Glove representation based model even without FOL constraints (M2) is comparable to SOTA method on both the datasets. 

\subsection{Qualitative Evaluation}
\begin{figure}[tb]
    \centering
    \resizebox{\linewidth}{6cm} {
    \includegraphics[width=\linewidth, keepaspectratio]{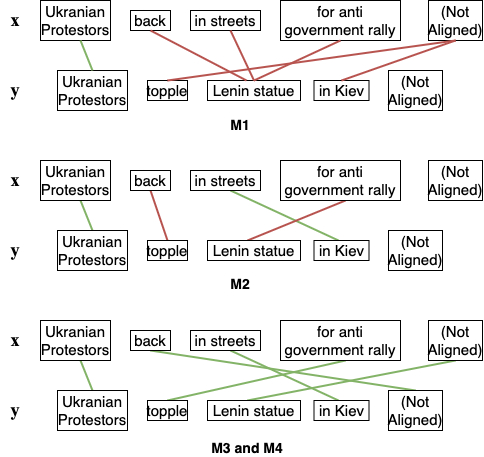}
    }
    \caption{(Best viewed in color) Relative merit of M3 \& M4 over M1 \& M2. The correct alignments according to ground truth are shown in green and the incorrect ones are shown in red. }
    \label{fig:qual_ex_1}
\end{figure}
We illustrate the qualitative merits of different components of our model on two examples by investigating the alignment produced by the four modelling configurations we introduced in Section~\ref{subsection:baselines}. Figure~\ref{fig:qual_ex_1} shows 
that modelling bidirectionality is important as is evident from large number of alignment errors made by M1 in comparison to M2. However, using BERT based contextual embeddings for chunks in the sentences leads to improved representations for chunks and both M3 and M4 correctly align all the chunks.  

\begin{figure}[tb]
    \centering
    \includegraphics[width=\linewidth, keepaspectratio]{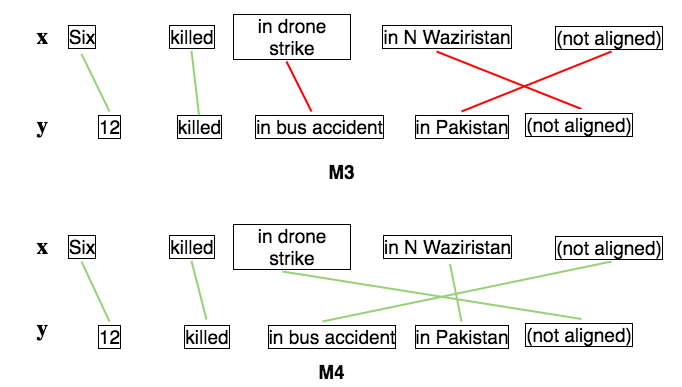}
    \caption{(Best viewed in color) Relative merit of M4 over M3. The correct alignments according to ground truth are shown in green and the incorrect ones are shown in red.}
    \label{fig:qual_ex_2}
\end{figure}

Figure~\ref{fig:qual_ex_2} shows the effect of including real-world knowledge in the form of FOL statements in the model. The chunks pair ``in N Waziristan'' and ``in Pakistan'' are aligned by the FOL statements retrieved from ConceptNet. This is because ConceptNet holds the real-world knowledge that ``Waziristan'' is a part of ``Pakistan'' and relates them. BERT embeddings alone may not encode enough information to relate ``Waziristan'' and ``Pakistan'' and M3 fails to align both these chunks correctly, while M4 owing to the activated FOL constraint gives the correct alignment.

\subsection{Cross-domain Experiments}
\label{subsec:cross-domain}
While the proposed approach achieves significant improvements over the previous baselines, a natural question arises -- how would this approach be applicable on a new sentence similarity dataset to provide interpretability, when no chunk alignments are available to train the model in the target domain? To answer this, we attempted a cross-domain experiment, where we train on headlines, test on images, and vice versa (i.e., do not utilize training examples from the target domain). Remarkably, even in this setting, using our best model M4, we achieve F1 scores of 96.16 and 94.80 on headlines and images datasets, respectively, outperforming the previous SOTA results. Note that no hyperparameter tuning was performed on the target domain.

\section{Conclusion}
We propose a novel pointer network for the task of interpretable sentence similarity along with logic constraints based on ConceptNet and syntactic knowledge. Experiments over benchmark datasets show a large performance improvement for the alignment task, even in the cross-domain setting, proving the general applicability of the proposed approach.

It was encouraging to see that the logical constraints imposed using external knowledge helped the model performance, and it would be interesting to check how the whole framework can be employed to improve performance for the sentence similarity task, while providing interpretability and explanation for the model decision.

\bibliographystyle{named}
\bibliography{ijcai20}

\end{document}